\documentclass[12pt]{article}

\usepackage{amsmath,amsthm,amsfonts,amssymb}
\usepackage{bm}

\usepackage{times}

\usepackage[margin=1in]{geometry}
\usepackage{setspace}
\usepackage{subcaption}

\usepackage{graphicx}
\usepackage{xcolor}
\usepackage{tikz}
\usetikzlibrary{automata, positioning, arrows, chains}
\usepackage[linguistics]{forest}
\usepackage{float}

\usepackage{booktabs}
\usepackage{multirow}
\usepackage{array}
\usepackage{caption}
\usepackage{tablefootnote}

\usepackage{algorithm}
\usepackage{algorithmic}

\usepackage{enumitem}

\usepackage{listings}
\usepackage{mdframed}

\usepackage{natbib}
\usepackage[hidelinks]{hyperref}
\usepackage{cleveref}
\usepackage{url}

\theoremstyle{plain}

\theoremstyle{definition}

\numberwithin{equation}{section}
\DeclareMathOperator*{\argmax}{arg\,max}

\newlength\myindent
\begin{document}

\title{Interpretable Machine Learning–Derived Spectral Indices for Vegetation Monitoring}

\author{
\begin{minipage}{0.85\textwidth}
\centering
Ali Lotfi$^{1,2}$,
Adam Carter$^{3}$,
Thuan Ha$^{1,2}$,
Mohammad Meysami$^{4}$,
Kwabena Nketia$^{1,2}$,
Steve Shirtliffe$^{1,2}$ \\[6pt]
\footnotesize $^{1}$Nutrien Centre for Sustainable and Digital Agriculture, University of Saskatchewan, Saskatoon, SK, Canada\\
$^{2}$Department of Plant Sciences, University of Saskatchewan, Saskatoon, SK, Canada\\
$^{3}$Crop Development Centre, Department of Plant Sciences, University of Saskatchewan, Saskatoon, SK, Canada\\
\footnotesize $^{4}$Department of Mathematics, The University of Tulsa, Tulsa, OK, USA
\end{minipage}
}
\date{}
\maketitle

\begin{abstract}
Spectral indices such as NDVI have driven vegetation monitoring for decades, yet their design remains largely manual and ad hoc. Their usefulness stems not only from their empirical performance, but also from algebraic forms that remain compact and biologically interpretable. However, the space of possible algebraic expressions relating spectral bands is effectively infinite, making systematic search impractical without structural constraints. We introduce the Spectral Feature Polynomial (SFP) framework, a general pipeline that automatically discovers compact, interpretable spectral indices from labeled multispectral imagery. SFP constructs a library of ratio-based spectral features that inherit illumination invariance by construction. It then applies cross-validated feature selection and continuous coefficient optimization to produce a single closed-form equation per task—transparent to domain experts and deployable on any remote sensing platform without requiring standardization statistics. We validate the framework on two agricultural applications. For Kochia (\textit{Bassia scoparia}) detection in Sentinel-2 imagery near Lucky Lake of Saskatchewan over three growing seasons, the same two-term equation emerged in 44 of 46 independent cross-validation folds, achieving 98.6\% mean accuracy, more than 4 percentage points above the best established index under year-held-out evaluation. For wheat plant classification from UAV multispectral imagery, stage-specific indices achieved 99.5\%, 97.2\%, and 93.5\% across three growth stages, compared to 78\% or below for the best established index at late season when NIR-based contrasts lose discriminatory power as wheat senesces. In both applications, SFP yielded a single transparent equation that generalized across held-out regions and outperformed established indices.
\end{abstract}

\section*{Introduction}

Remote sensing enables consistent vegetation monitoring over regions that are impractical to survey in the field \cite{xue2017significant}. A cornerstone of multispectral analysis is the normalized-difference family of ratio indices, which is computationally simple, relatively illumination-robust, and transferable across sensors \cite{bannari1995review}.

The Normalized Difference Vegetation Index (NDVI) \citep{rouse1973monitoring} is the canonical normalized-difference index, leveraging red absorption and near-infrared reflectance of healthy vegetation to quantify greenness \cite{tucker1979red}.

The success of NDVI has led to many additional spectral indices designed to address specific limitations or exploit additional spectral features. The Soil Adjusted Vegetation Index (SAVI) introduces a soil brightness correction to reduce background soil influence in sparse vegetation \cite{huete1988soil}. The Enhanced Vegetation Index (EVI) further incorporates blue reflectance to correct atmospheric effects and reduce saturation in areas with dense biomass \cite{huete1994development}.

Sensors with dedicated red-edge bands—particularly the Sentinel-2 MSI—extend this paradigm to chlorophyll-sensitive contrasts. Red-edge indices such as NDRE are more sensitive to chlorophyll content and nitrogen status \cite{sonobe2018crop}, improve separation of vegetation species at fine taxonomic levels \cite{immitzer2016first, kang2021crop}, and enable reliable weed detection in cereal fields with producer accuracies above 88\% \cite{mudereri2019comparative}.

Despite these advances, some invasive species remain difficult to identify. Kochia (\textit{Bassia scoparia}) is particularly challenging because it often occurs within crop fields and exhibits spectral responses similar to surrounding vegetation. Hyperspectral imagery studies have reported classification accuracies of 67--80\% when separating herbicide-resistant Kochia biotypes using support vector machine classifiers \cite{nugent2018discrimination, scherrer2019hyperspectral}. More recent work using attention-based convolutional neural networks has pushed accuracy above 99\% when separating Kochia from sugarbeet under field conditions \cite{mensah2025detection}. While powerful, these approaches often require costly hyperspectral sensors and complex computational pipelines, limiting their practicality for routine regional monitoring.

Deep learning models can achieve high classification accuracy in remote sensing, but their black-box nature often limits practical adoption. Practitioners in earth observation—such as farmers, agronomists, and policy makers—need to understand why a model flags a pixel before acting on that prediction \cite{reichstein2019deep, murdoch2019definitions, mcgovern2019making}. Interpretable spectral functions provide this transparency directly by linking predictions to known biophysical processes \cite{de2000classification}. In addition, parsimonious models often generalize better across sensors and deployment conditions, require fewer computational resources, and can be implemented on virtually any remote sensing platform \cite{duro2012comparison, laaha2007national}.

New spectral indices are typically designed through expert judgment and trial and error, with researchers writing candidate formulas from spectral intuition and testing them against ground truth \cite{basso2019seasonal, tiedeman2022field}. This approach is slow and unlikely to find optimal solutions given the effectively infinite space of possible spectral formulas: even a ten-band sensor yields 45 normalized differences, over 120 three-band combinations, and many more formulas involving products, powers, and nested ratios. With 13 bands on Sentinel-2 MSI, the space of candidate indices becomes effectively intractable \cite{montero2023standardized}. The problem is compounded by context dependence: an index tuned for crop stress may perform poorly at weed detection, and a formula calibrated for one sensor may not transfer to another. This makes automated, data-driven index discovery both necessary and practically useful.

Feature selection methods from machine learning provide another route for identifying the small subsets of spectral features that carry the most discriminative information \cite{guyon2003introduction}. Building on this idea, we introduce the \emph{Spectral Feature Polynomial} (SFP) framework for discovering compact, interpretable spectral indices from multispectral imagery. The framework constructs a structured feature space of ratio-based spectral functions and applies automated feature selection with coefficient optimization to identify a single deployable equation for a given task. We evaluate the approach on two agricultural applications using Sentinel-2 satellite imagery and UAV multispectral imagery. Section~\ref{sec:methods} describes the SFP framework, including the spectral basis families, the selection protocol, and coefficient optimization. Section~\ref{sec:satellite_results} presents the Kochia detection application using Sentinel-2 imagery. Section~\ref{sec:uav_results} presents the wheat plant classification application using UAV multispectral imagery. Section~\ref{discussion} discusses the framework in relation to existing automated index discovery methods, and Section~\ref{conclusion} concludes.

\section{Methods}\label{sec:methods}

Figure~\ref{fig:pipeline} and Algorithm~\ref{alg:pipeline} summarize the SFP pipeline; we describe each component below.

\begin{figure}[htbp]
\centering
\begin{tikzpicture}[
    node distance=0.45cm,
    box/.style={
        rectangle, rounded corners=3pt,
        draw=black, fill=gray!10,
        text width=2.0cm, align=center,
        minimum height=1.0cm, font=\small
    },
    subbox/.style={
        rectangle, rounded corners=2pt,
        draw=gray, fill=white,
        text width=2.0cm, align=center,
        minimum height=0.7cm, font=\scriptsize,
        text=gray
    },
    arrow/.style={->, >=stealth, thick}
]

% Main boxes
\node[box] (input)  {Multispectral\\Bands};
\node[box, right=of input]  (basis)  {Basis\\Construction};
\node[box, right=of basis]  (poly)   {Polynomial\\Expansion};
\node[box, right=of poly]   (select) {Feature\\Selection};
\node[box, right=of select] (coeff)  {Coefficient\\Optimization};
\node[box, right=of coeff, fill=black!80, text=white]  (output) {Interpretable\\Index};

% Sub-annotation boxes
\node[subbox, below=of basis]  (s1) {13 families\\ND, ND3, NCurv\\+ 10 extended};
\node[subbox, below=of poly]   (s2) {degree-2\\products \&\\squares};
\node[subbox, below=of select] (s3) {ANOVA pre-filter\\SelectKBest / RFE\\consensus across folds};
\node[subbox, below=of coeff]  (s4) {Fisher grid search\\fine zoom\\Nelder-Mead};

% Arrows between main boxes
\draw[arrow] (input)  -- (basis);
\draw[arrow] (basis)  -- (poly);
\draw[arrow] (poly)   -- (select);
\draw[arrow] (select) -- (coeff);
\draw[arrow] (coeff)  -- (output);

% Dotted lines to sub-annotations
\draw[dotted, gray] (basis)  -- (s1);
\draw[dotted, gray] (poly)   -- (s2);
\draw[dotted, gray] (select) -- (s3);
\draw[dotted, gray] (coeff)  -- (s4);

\end{tikzpicture}
\caption{Overview of the SFP pipeline: basis construction, degree-2 expansion, cross-validated selection, and coefficient optimization yield a single deployable index.}
\label{fig:pipeline}
\end{figure}

\textbf{Basis families.} Let a multispectral sensor provide $n$ reflectance bands $b_1,\ldots,b_n$ with $b_i \ge 0$. SFP builds on normalized differences, which are invariant to multiplicative scaling~\cite{unger2007introductory}, and extends them to three core families. The two-band normalized difference, for any pair $b_i, b_j$ with $i \neq j$, is defined as
\begin{equation}
\mathrm{ND}(b_i, b_j) = \frac{b_i - b_j}{b_i + b_j + \varepsilon}
\end{equation}
where $\varepsilon = 10^{-10}$ ensures numerical stability. This yields $\binom{n}{2}$ pairwise contrast features. To capture three-band spectral relationships we introduce the three-band normalized difference:
\begin{equation}
\mathrm{ND3}(s_1 b_i, s_2 b_j, s_3 b_k) = \frac{s_1 b_i + s_2 b_j + s_3 b_k}{b_i + b_j + b_k + \varepsilon}
\end{equation}
where $(s_1, s_2, s_3) \in \{(+,+,-), (+,-,+), (-,+,+)\}$ are sign patterns with exactly one negative term, producing $3\binom{n}{3}$ features. The third core family, the normalized curvature, captures spectral curvature at a middle band relative to its neighbors:
\begin{equation}
\mathrm{NCurv}(b_i, b_j, b_k) = \frac{b_i - 2b_j + b_k}{b_i + 2b_j + b_k + \varepsilon}
\end{equation}
contributing $\binom{n}{3}$ features. The numerator is a discrete second derivative at $b_j$ (near zero: locally linear; sign indicates convexity or concavity), making this family useful for detecting red-edge inflection behavior. All three core families are illumination-invariant and bounded in $(-1,1)$ for non-negative reflectance values, keeping downstream features numerically stable.

The framework also supports ten additional basis families that broaden the vocabulary of ratio-based spectral contrasts (Table~\ref{tab:basis_families}). These fall into four conceptual groups. Pairwise ratio families include SR, a bounded $\tanh$-transformed generalization of the classic Simple Ratio vegetation index~\cite{jordan1969derivation}, and BandRatio, which compares each band against the running spectral maximum. Band-fraction families measure fractional contributions to multi-band sums, inspired by spectral abundance and mixture-fraction concepts in hyperspectral analysis. Asymmetric sum-ratio families (SR2 and SR3) generalize the symmetric numerator--denominator structure of normalized differences to unequal cardinalities, motivated by multi-band indices such as EVI~\cite{huete2002overview}. Finally, spectral height families (HSurr and HAbs) measure the prominence of a band's peak above its spectral neighborhood, analogous to band depth and peak prominence in absorption spectroscopy, together with a global shape summary (SpecCV) that captures overall spectral variability.

\begin{table}[htbp]
\centering
\caption{Additional basis families supported by the SFP framework, organized by conceptual group. $\bar{b}_{-i}$ denotes the mean of all bands except $b_i$; $\varepsilon = 10^{-10}$.}
\label{tab:basis_families}
\begin{tabular}{llp{5.2cm}p{4.0cm}}
\toprule
Group & Family & Formula & Spectral interpretation \\
\midrule
\multirow{2}{*}{Pairwise ratio}
 & SR        & $\tanh(b_i/b_j - 1)$
             & Bounded simple ratio; captures pairwise band contrasts \\[4pt]
 & BandRatio & $(b_i - \max_{j\neq i} b_j)\,/\,(b_i + \max_{j\neq i} b_j + \varepsilon)$
             & Normalized contrast of each band against the current spectral maximum \\
\midrule
\multirow{4}{*}{Band fraction}
 & BOP   & $b_i\,/\,(b_j + b_k + \varepsilon)$
         & Fractional contribution of one band to a two-band denominator \\[4pt]
 & BOT   & $b_i\,/\,(b_j + b_k + b_l + \varepsilon)$
         & Fractional contribution of one band to a three-band denominator \\[4pt]
 & BOA   & $b_i\,/\,(\sum_{j\neq i} b_j + \varepsilon)$
         & Fraction of a band relative to all other bands combined \\[4pt]
 & BFrac & $b_i\,/\,(\sum_j b_j + \varepsilon)$
         & Fraction of a band relative to the total spectral sum \\
\midrule
\multirow{2}{*}{Asymmetric ratio}
 & SR2 & $(b_i + b_j)\,/\,(b_k + b_l + \varepsilon)$
       & Two-band sum divided by two-band sum; generalizes ND to asymmetric cardinality \\[4pt]
 & SR3 & $(b_i + b_j + b_k)\,/\,(b_l + \varepsilon)$
       & Three-band sum divided by a single reference band \\
\midrule
\multirow{3}{*}{Height / shape}
 & HSurr  & $\tanh\!\bigl((b_i - \max_{j\neq i} b_j)\,/\,\bar{b}_{-i}\bigr)$
           & Scale-invariant peak prominence; illumination robust \\[2pt]
 & HAbs   & $\tanh(b_i - \max_{j\neq i} b_j)$
           & Absolute peak prominence; illumination sensitive \\[2pt]
 & SpecCV & $\sigma_{\mathbf{b}}\,/\,\mu_{\mathbf{b}}$
           & Coefficient of variation across all bands \\
\bottomrule
\end{tabular}
\end{table}

Which families are included depends on sensor band count and available training data. For sensors with many bands (e.g., Sentinel-2), the three core families already produce a sufficiently rich degree-2 space. For lower-band sensors such as the five-band MicaSense RedEdge, all 13 families are included to keep the search space expressive. Most families share a common structure—ratios whose numerator and denominator are linear combinations of band values—enabling the coefficient optimization described later.

\textbf{Polynomial expansion.} To capture nonlinear interactions between spectral patterns, we construct a degree-$d$ polynomial expansion of the basis terms. Let $\mathcal{B}$ denote the union of all selected basis functions. The polynomial feature space of degree $d$ is
\begin{equation}
\Phi = \left\{ \prod_{j=1}^{|\mathcal{B}|} \phi_j^{k_j} \;:\; \phi_j \in \mathcal{B},\; k_j \in \mathbb{Z}_{\geq 0},\; \sum_j k_j \leq d \right\}
\end{equation}
We use $d = 2$, producing three types of features: the basis terms themselves, their squares, and pairwise products of distinct basis terms. Degree-2 was chosen to balance expressiveness with tractability: it captures multiplicative interactions between spectral contrasts while keeping the feature space computationally manageable. For the Sentinel-2 application ($n = 9$ bands, three core families: ND, ND3, NCurv), the basis contains $|\mathcal{B}| = 372$ terms (36 ND, 252 ND3, 84 NCurv), yielding a degree-2 feature space of $|\Phi| = 69{,}750$. For the UAV application ($n = 5$ bands, all 13 families), the basis contains $|\mathcal{B}| = 197$ terms, yielding $|\Phi| = 19{,}700$ features.

\textbf{Feature selection.} From this space, we seek the smallest subset $\mathcal{S} \subseteq \Phi$ that achieves high classification accuracy with a linear support vector machine (SVM). Parsimony is preferred: a single-feature model is favored over a slightly more accurate multi-feature predictor, because it yields a deployable equation rather than a black-box model.

The selection proceeds in three stages. First, a global ANOVA pre-filter ranks all polynomial features by their univariate F-statistic~\cite{guyon2003introduction} and retains the top $p$ (typically $p = 500$--$2{,}000$), reducing the search space by one to two orders of magnitude. Second, for each cross-validation fold, two complementary methods are applied independently: SelectKBest, which ranks the remaining features by F-statistic and retains the top $k$, and Recursive Feature Elimination (RFE)~\cite{guyon2002gene}, which iteratively removes the least important feature based on linear SVM coefficient magnitude. For each target feature count $k = 1, \ldots, k_{\max}$, the method achieving higher test accuracy on that fold is retained.

Third, we identify the feature count at which accuracy plateaus and select the index that appears most frequently across folds at that count. Rather than relying on a single train/test split, we require the same feature to emerge independently across folds that differ in temporal composition, spatial extent, or both. Once the top index is identified, we tune the SVM regularization parameter $C$ over a grid of 20 values from $10^{-4}$ to $10^{3}$ using the same cross-validation folds, selecting the value that maximizes mean accuracy. In the rare case where two features appear with equal frequency across folds, the simpler expression — defined as the one involving fewer distinct bands — is preferred, consistent with the parsimony objective.

\textbf{Coefficient optimization.} The selection pipeline discovers the structural form of the index: which bands appear and in which functional arrangement. However, the initial weights within each basis term are fixed at their default values (unit weights for ND, ND3, and ratio families; weights $[1, 2, 1]$ for NCurv). We introduce a post-selection optimization that learns continuous coefficients while preserving the normalized difference structure. Each term is expressed in the unified weighted form:
\begin{equation}
\label{eq:weighted_term}
T(\mathbf{w}) = \frac{\sum_i s_i \, w_i \, b_i}{\sum_i w_i \, b_i + \varepsilon}
\end{equation}
where $s_i \in \{-1, +1\}$ are the sign patterns from the original basis function and $w_i > 0$ are continuous weights. The first weight in each term is fixed at 1.0 to eliminate scale redundancy, and a product index is formed by multiplying $m$ such terms.

The optimization proceeds in four phases. First, a coarse grid search evaluates Fisher's discriminant ratio
\begin{equation}
J(\mathbf{w}) = \frac{(\mu_+ - \mu_-)^2}{\sigma_+^2 + \sigma_-^2}
\end{equation}
across the weight space, where $\mu_\pm$ and $\sigma_\pm^2$ are the class-conditional means and variances of the index. Because Fisher's ratio is computed from the data without involving a classifier, this search can use all available samples without risk of cross-validation leakage. Second, a finer grid search around the top candidates narrows the search region. Third, the best candidates from these grid searches are evaluated through the full cross-validation protocol to identify weights that generalize well. Fourth, Nelder-Mead optimization~\cite{nelder1965simplex} is applied to polish the top candidates, directly maximizing cross-validation accuracy.

The resulting equation runs directly on reflectance values and can be deployed in any remote sensing platform, including Google Earth Engine, without requiring access to training data or feature standardization.

\begin{algorithm}[htbp]
\caption{ndindex: Spectral Index Discovery Pipeline}
\label{alg:pipeline}
\begin{algorithmic}[1]
\REQUIRE Band reflectances $\mathbf{B} \in \mathbb{R}^{N \times n}$, labels $\mathbf{y}$, cross-validation folds $\mathcal{F}$
\ENSURE Optimized index equation and decision threshold

\medskip
\STATE \textit{// Basis construction and polynomial expansion}
\STATE Generate basis $\mathcal{B}$ from selected families
\STATE Expand to degree-$d$ polynomial space $\Phi$ (products and squares of $\mathcal{B}$)

\medskip
\STATE \textit{// Global pre-filter}
\STATE Rank all $|\Phi|$ features by ANOVA F-statistic; retain top $p$

\medskip
\STATE \textit{// Per-fold feature selection}
\FOR{each fold $f \in \mathcal{F}$}
    \FOR{$k = 1$ to $k_{\max}$}
        \STATE $\mathcal{S}_{\text{KB}} \gets$ top $k$ features by F-statistic on training set
        \STATE $\mathcal{S}_{\text{RFE}} \gets$ top $k$ features by recursive elimination (linear SVM)
        \STATE Retain whichever method achieves higher test accuracy
    \ENDFOR
\ENDFOR

\medskip
\STATE \textit{// Consensus and hyperparameter tuning}
\STATE $k^* \gets$ feature count where accuracy plateaus
\STATE $\phi^* \gets$ most frequently selected feature at $k^*$ across all folds
\STATE $C^* \gets \argmax_{C} \frac{1}{|\mathcal{F}|} \sum_{f \in \mathcal{F}} \text{Acc}_f(\phi^*, C)$

\medskip
\STATE \textit{// Coefficient optimization}
\STATE Parse $\phi^*$ into terms with default weights $\mathbf{w}_0$
\STATE Coarse grid search: maximize Fisher's ratio $J(\mathbf{w})$ over weight space
\STATE Fine grid zoom: refine around top candidates
\STATE Evaluate top candidates through full CV with SVM($C^*$)
\STATE Nelder-Mead polish: locally maximize CV accuracy
\STATE $\mathbf{w}^* \gets$ weights with highest CV accuracy

\medskip
\STATE \textit{// Deployment}
\STATE Compute threshold from SVM decision boundary on $\phi^*(\mathbf{w}^*)$
\RETURN Index equation $\phi^*(\mathbf{w}^*)$ and threshold
\end{algorithmic}
\end{algorithm}

\section{Application to Satellite Imaging}
\label{sec:satellite_results}

We applied the SFP framework to detect Kochia (\textit{Bassia scoparia}) in Sentinel-2 imagery collected near Lucky Lake, Saskatchewan, Canada (51.06° N, 107.27° W) over three growing seasons (2022–2024). Candidate fields were first identified using UAV imagery. Sentinel-2 scenes acquired on August 14 of each year were used, corresponding to the late growing season when spectral contrast between Kochia and crop canopies is strongest. Ground truth was obtained by delineating interior polygons over known Kochia and crop areas while excluding field boundaries to avoid spatial leakage. Pixel values extracted at these locations produced 2,196 labeled samples (Table~\ref{tab:dataset_summary}).

The imagery contained ten Sentinel-2 bands: Blue (B2, 490\,nm), Green (B3, 560\,nm), Red (B4, 665\,nm), three red-edge bands (B5, 705\,nm; B6, 740\,nm; B7, 783\,nm), NIR (B8, 842\,nm), narrow NIR (B8A, 865\,nm), SWIR-1 (B11, 1610\,nm), and SWIR-2 (B12, 2190\,nm). Band B8A was excluded due to strong redundancy with B8, leaving nine bands for index discovery.

\begin{table}[htbp]
\centering
\caption{Pixel sample composition after quality control.}
\label{tab:dataset_summary}
\begin{tabular}{lcccc}
\toprule
 & 2022 & 2023 & 2024 & Total \\
\midrule
Kochia & 319 & 555 & 324 & 1{,}198 \\
Crop   & 268 & 424 & 306 & 998 \\
\midrule
Total  & 587 & 979 & 630 & 2{,}196 \\
\bottomrule
\end{tabular}
\end{table}

From this feature space the framework converged on a single equation, the \textit{Kochia Detection Index} (KDI):

\begin{align}
\label{eq:kdi}
\text{KDI} &= \frac{-\text{B7} + w_1\,\text{B8} + w_2\,\text{B11}}{\text{B7} + w_1\,\text{B8} + w_2\,\text{B11}} \;\times\; \frac{\text{B5} - w_3\,\text{B7} + w_4\,\text{B8}}{\text{B5} + w_3\,\text{B7} + w_4\,\text{B8}} \notag \\
&= \frac{-\text{RE3} + w_1\,\text{NIR} + w_2\,\text{SWIR-1}}{\text{RE3} + w_1\,\text{NIR} + w_2\,\text{SWIR-1}} \;\times\; \frac{\text{RE1} - w_3\,\text{RE3} + w_4\,\text{NIR}}{\text{RE1} + w_3\,\text{RE3} + w_4\,\text{NIR}}
\end{align}

\noindent where B5 (705\,nm), B7 (783\,nm), B8 (842\,nm), and B11 (1610\,nm) are Sentinel-2 reflectance bands and the optimized coefficients are $w_1 = 1.09$, $w_2 = 0.37$, $w_3 = 1.99$, $w_4 = 0.80$. The first term captures the contrast between the red-edge shoulder and the combined NIR/SWIR response, reflecting structural differences between Kochia and cereal crop canopies. The second term measures spectral curvature across the red-edge, where Kochia exhibits a characteristically steep transition. The low optimized weight on B11 ($w_2 = 0.37$) suggests the SWIR band provides a useful but secondary correction to the primary red-edge/NIR discrimination.

This index was independently selected as the top feature in 44 of 46 cross-validation folds (96\%), despite each fold performing feature selection on a different subset of the data, yet the same equation emerges whether the holdout is temporal (a different year), spatial (a different geographic block), or both. The two folds that selected a different index chose structurally related variants differing by a single band substitution.

We evaluated the KDI across four cross-validation strategies of increasing 
stringency (Table~\ref{tab:cv_results}): random stratified 10-fold, 
year-held-out (training on two seasons, testing on the third), spatial 
block holdout (a $3 \times 3$ geographic grid, each block held out once), 
and spatio-temporal block holdout. Each spatio-temporal fold withholds all 
samples from one spatial block in one specific year, training on the 
remaining blocks from that year combined with all samples from the other 
two years. With a $3 \times 3$ spatial grid and three years, this yields up to 27 folds. Of these, 24 contained sufficient samples for evaluation.
Together, these 46 folds test whether the index generalizes across temporal 
variation, spatial variation, and the combination of both.

\begin{table}[htbp]
\centering
\caption{Classification accuracy (\%) of the KDI with a linear SVM ($C = 0.5$) across four cross-validation strategies.}
\label{tab:cv_results}
\begin{tabular}{lcccc}
\toprule
Strategy & Folds & Mean & Median & Min \\
\midrule
Random 10-fold          & 10 & 98.7 & 98.9 & 97.3 \\
Year-held-out           &  3 & 98.6 & 99.4 & 96.3 \\
Spatial block           &  9 & 98.2 & 99.1 & 93.6 \\
Spatio-temporal block   & 24 & 98.7 & 100.0 & 86.7 \\
\midrule
\textbf{Overall}        & \textbf{46} & \textbf{98.6} & \textbf{99.5} & \textbf{86.7} \\
\bottomrule
\end{tabular}
\end{table}

Mean accuracy exceeded 98\% in every strategy. Coefficient optimization improved accuracy by 1.1--1.5 percentage points across all strategies (Table~\ref{tab:coeff_optimization}). Across all 46 folds, 42 exceeded 95\% accuracy and 21 achieved 100\%, with the few underperforming folds concentrated in spatio-temporal blocks from 2024.

\begin{table}[htbp]
\centering
\caption{Effect of coefficient optimization on mean accuracy (\%).}
\label{tab:coeff_optimization}
\begin{tabular}{lccc}
\toprule
Strategy & Default & Optimized & $\Delta$ \\
\midrule
Random 10-fold          & 97.5 & 98.7 & +1.2 \\
Year-held-out           & 97.1 & 98.6 & +1.5 \\
Spatial block           & 97.1 & 98.2 & +1.1 \\
Spatio-temporal block   & 97.3 & 98.7 & +1.3 \\
\bottomrule
\end{tabular}
\end{table}

\begin{figure}[htbp]
    \centering
    \includegraphics[width=\textwidth]{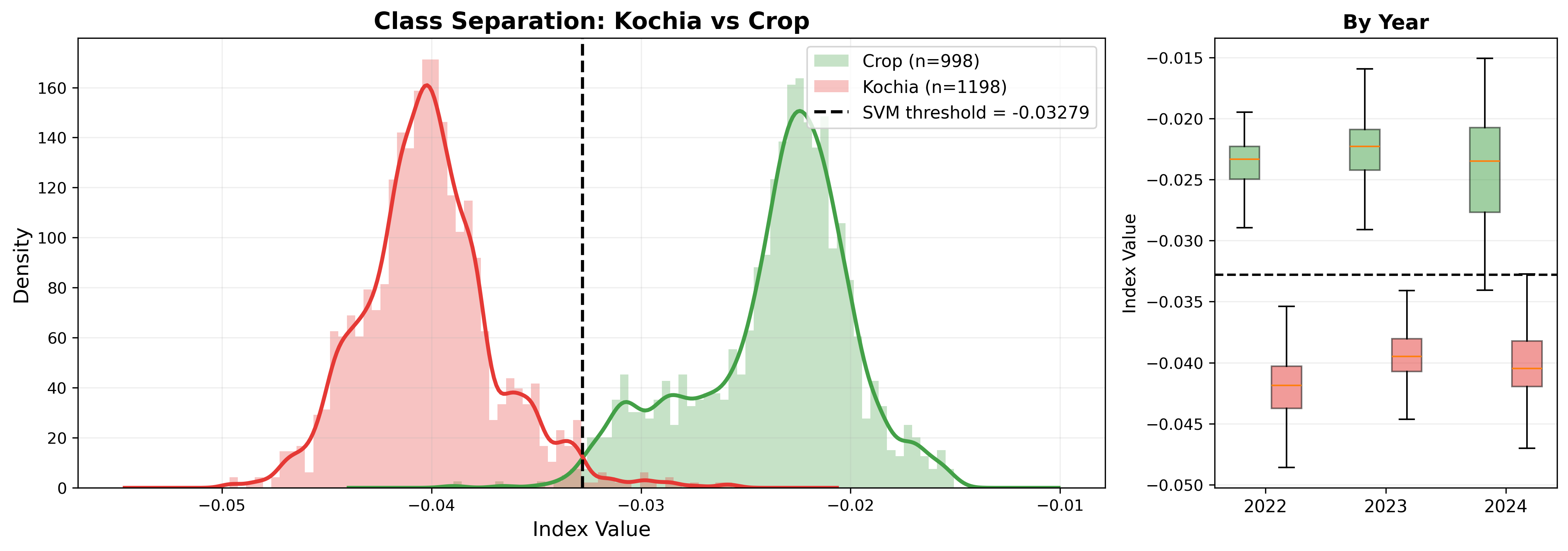}
    \caption{Distribution of KDI values for Kochia and Crop, with the SVM decision threshold in the region of minimal overlap.}
    \label{fig:class_separation}
\end{figure}

The two classes form clearly separated distributions with minimal overlap (Figure~\ref{fig:class_separation}), and this separation is maintained across all three years. The 2024 season shows a narrower gap, consistent with its lower year-held-out accuracy (96.3\%), though the index still performs well overall. One spatio-temporal fold in 2024 achieved 86.7\%. This reflects genuine spectral similarity between classes at that location: no candidate index, including NDVI and NDRE, exceeded 57\% on that fold, confirming that the KDI substantially outperformed alternatives even under the most challenging conditions.

\begin{figure}[htbp]
    \centering
    \includegraphics[width=1\textwidth]{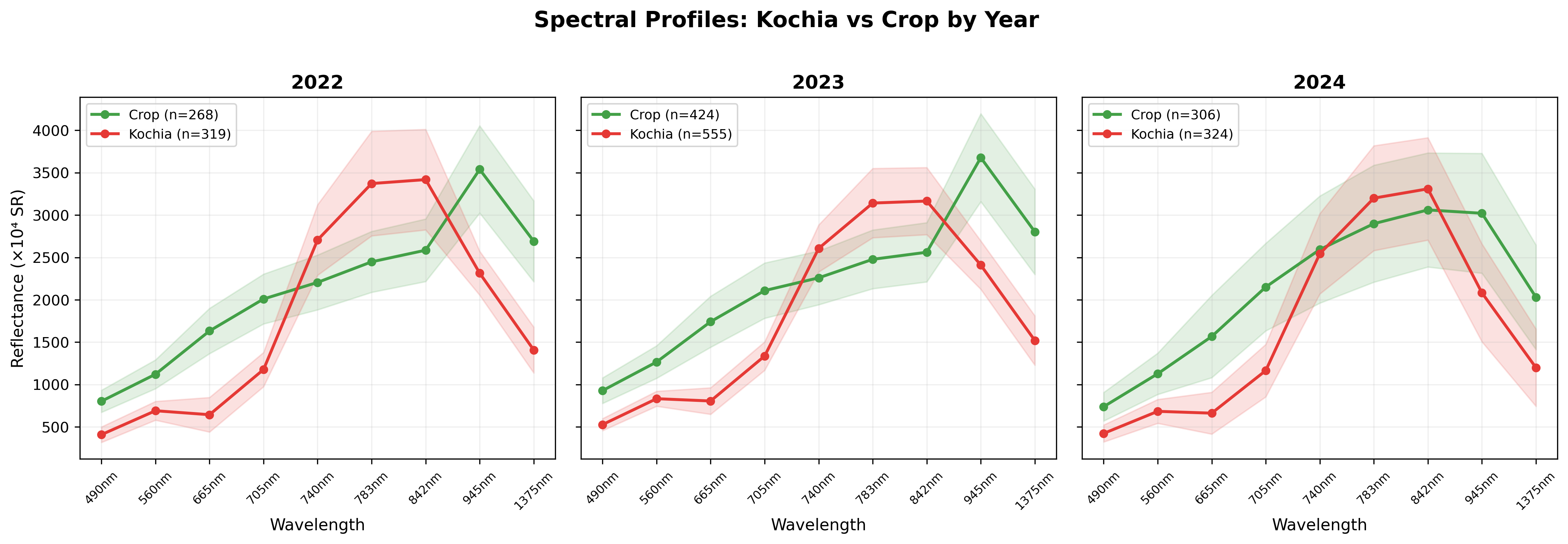}
    \caption{Mean reflectance ($\pm$\,1 SD) for Kochia and Crop across Sentinel-2 bands, by year. The red-edge region (B5--B7) shows the most consistent class separation.}
    \label{fig:spectral_profiles}
\end{figure}

The spectral basis for this discrimination is illustrated in Figure~\ref{fig:spectral_profiles}. The largest and most stable class differences occur in the red-edge (705--783\,nm), where Kochia's distinct canopy architecture produces consistently higher reflectance than cereal crops across all three years.

\begin{figure}[htbp]
    \centering
    \includegraphics[width=0.6\textwidth]{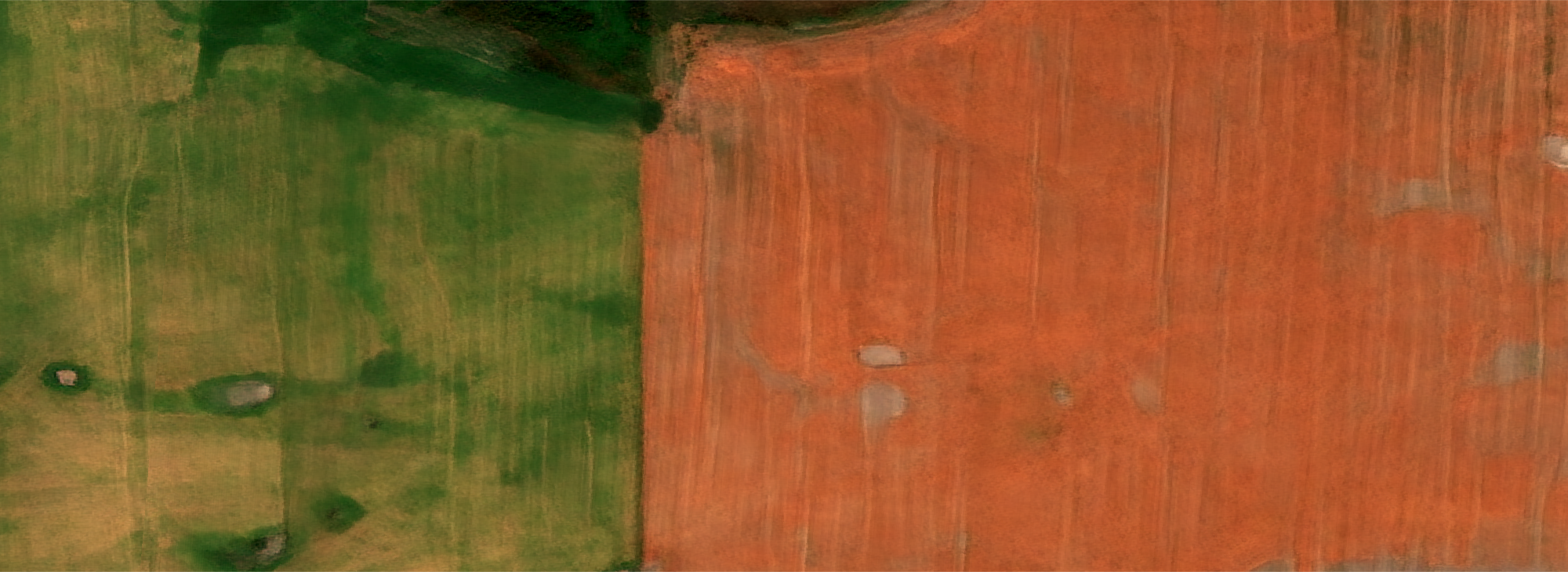}\\[4pt]
    \includegraphics[width=0.6\textwidth]{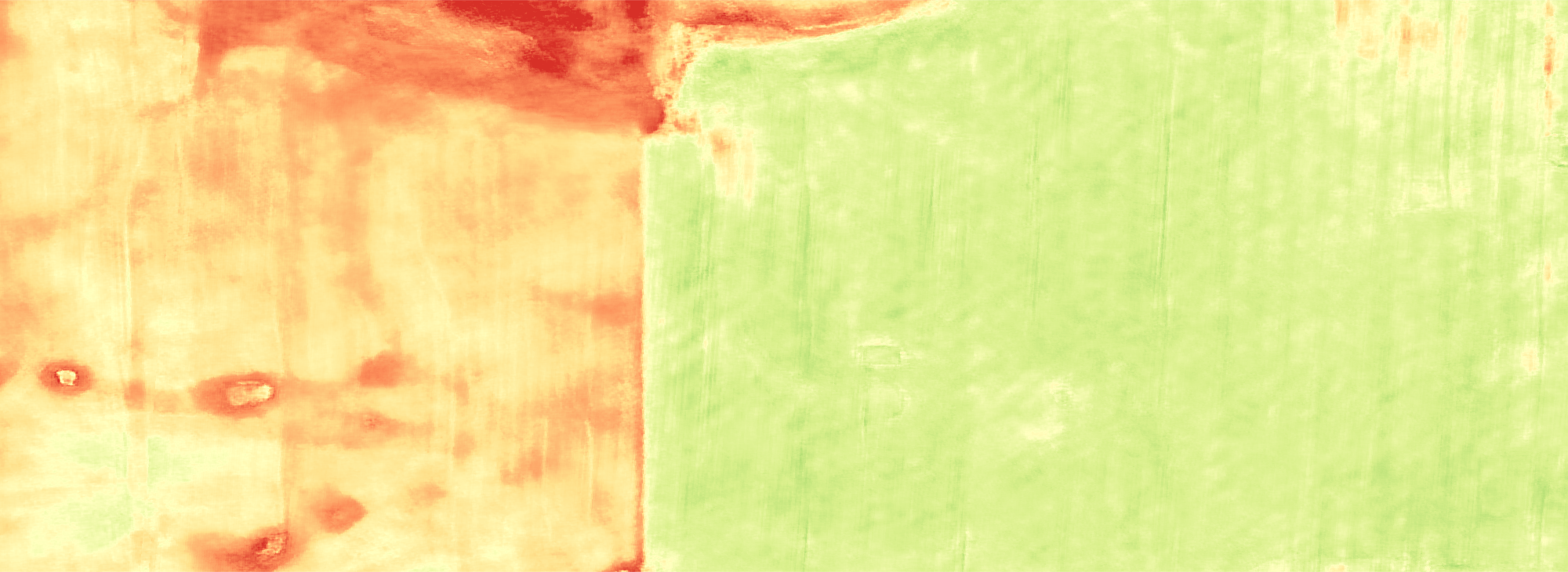}
    \caption{Sentinel-2 imagery acquired on August 14, 2024 over a mixed 
    Kochia--crop area at Lucky Lake, Saskatchewan. Top: true-color RGB 
    composite. Bottom: KDI values. The hard boundary between left and right reflects a genuine field boundary visible in the RGB image between a Kochia-infested area and an 
    adjacent crop field, not a cross-validation split. Kochia patches appear 
    in the left portion of the scene while crop fields dominate the right. 
    The KDI map reveals a clear spectral contrast between the two areas.}
    \label{fig:kdi_spatial}
\end{figure}

The spatial performance of the KDI is illustrated in Figure~\ref{fig:kdi_spatial}, which compares a true-color RGB composite with the corresponding KDI map over a mixed Kochia--crop area. While the two classes are visually similar in the RGB image, the KDI produces a clear spatial contrast between Kochia and crop fields.

We compared the KDI against six widely used vegetation indices, each evaluated with the same linear SVM classifier and cross-validation strategies (Table~\ref{tab:baseline_comparison}). The KDI consistently outperformed all baselines, with the largest margins under the year-held-out strategy, where temporal generalization is most challenging. NDRE and CIre, both of which incorporate red-edge information, were the strongest competitors, while broadband indices such as NDVI and GNDVI showed substantially lower accuracy.

\begin{table}[htbp]
\centering
\caption{Comparison of mean classification accuracy (\%) between the KDI and standard vegetation indices across four cross-validation strategies.}
\label{tab:baseline_comparison}
\begin{tabular}{lcccc}
\toprule
Index & Random & YHO & Spatial & ST \\
\midrule
\textbf{KDI} & \textbf{98.7} & \textbf{98.6} & \textbf{98.2} & \textbf{98.7} \\
NDRE  & 95.7 & 94.1 & 95.0 & 94.8 \\
CIre  & 95.9 & 93.8 & 95.1 & 95.0 \\
NDVI  & 94.8 & 92.4 & 94.0 & 93.4 \\
SAVI  & 94.8 & 92.4 & 94.0 & 93.4 \\
EVI   & 94.7 & 93.9 & 93.9 & 93.6 \\
GNDVI & 92.0 & 86.8 & 91.4 & 91.1 \\
\bottomrule
\end{tabular}
\end{table}

These results demonstrate that an automatically discovered index can reliably distinguish Kochia from crop across years, locations, and evaluation strategies using a single compact equation and a linear decision boundary.

\section{Application to UAV Imagery: Wheat Canopy Classification}
\label{sec:uav_results}

We applied the SFP framework to a second domain: detecting wheat canopy in UAV multispectral imagery acquired at ~15 m altitude (AGL) over a small-plot breeding trial at the University of Saskatchewan Seed Farm location with plot sizes of approximately 0.40 m by 0.24 m. This application differs fundamentally  from the satellite case in spatial resolution, sensor type, and classification challenge. At the centimeter-scale resolution of UAV imagery, canopy and background elements (non-wheat vegetation, soil, and survey markers) are spatially interleaved at the sub-plant level, and the spectral differences between wheat and background vegetation (including volunteer barley and occasional broadleaf weeds) shift substantially as the season progresses, posing a distinct discrimination challenge at each growth stage. The late-season window is therefore the critical test of any candidate index.

\subsection{Data}

Imagery was collected at three growth stages (Zadoks 39, 59, and 87) using a MicaSense RedEdge multispectral camera mounted on a UAV, providing five spectral bands: Blue ($b_1$, 475\,nm), Green ($b_2$, 560\,nm), Red ($b_3$, 668\,nm), Red-edge ($b_4$, 717\,nm), and NIR ($b_5$, 840\,nm). The three acquisition dates correspond to Growth Stage~1 (Zadoks 39; early canopy establishment), Growth Stage~2 (Zadoks 59; mid-season canopy closure), and Growth Stage~3 (Zadoks 87; late season, when wheat begins to senesce while other vegetation remains actively green, creating an inverted spectral ordering relative to earlier stages). Four classes were labelled: Canopy (wheat), non-wheat vegetation (which may include 
volunteer barley and occasional broadleaf weeds), Soil, and Ground Control Point (GCP) markers. The binary task is wheat canopy vs.\ non-wheat background (non-wheat vegetation, soil, and markers)

Ground truth samples were collected by drawing polygons over each class in the UAV imagery using a custom labelling tool, with pixel values extracted at each polygon location. A total of 6{,}000 pixel samples were collected per growth stage (18{,}000 overall), distributed across classes according to their field prevalence: 
2{,}700 Canopy, 2{,}400 non-wheat vegetation, 600 Soil, and 300 GCP samples per stage.

\subsection{Spatial Cross-Validation Design}

UAV imagery presents a challenge that differs from the satellite case: adjacent pixels share nearly identical reflectance values due to the very high spatial resolution, so random cross-validation splits place training and test pixels within centimeters of each other. This spatial autocorrelation inflates accuracy estimates and masks genuine generalization failures~\cite{roberts2017cross}. To obtain honest accuracy estimates, we partitioned each growth stage's samples into two spatially separated blocks using a horizontal split line across the field. All samples above the line were assigned to Block~1 and all samples below to Block~2. The two-fold evaluation trained on Block~1 and tested on Block~2, then reversed. A naive majority-class classifier achieves 55.0\% accuracy at each growth stage, confirming that the classification task is non-trivial despite the relatively balanced class distribution.

\subsection{Discovered Indices and Performance}

We term the discovered indices collectively the \textit{Wheat Canopy Index} (WCI), subscripted by growth stage to distinguish the three stage-specific formulations. The SFP framework was applied independently to each growth stage using all 13 basis families, searching a space of 19{,}700 candidate features. No coefficient optimization was applied at this stage; this is a deliberate conservative choice that demonstrates the framework's feature selection alone, without post-selection weight tuning, is sufficient to substantially outperform established indices. Because the indices are products of signed ratios, their absolute values may be negative; classification depends only on the learned decision threshold, so the canopy class simply occupies the highest portion of the index range. The identified indices are reported below; band notation follows the MicaSense channel ordering defined above.

\textbf{Growth Stage~1 index} (Zadoks 39 -- early season):
\begin{equation}
\label{eq:wci_t1}
\text{WCI}_{1} = -\frac{b_1 - b_3 + b_4}{b_1 + b_3 + b_4 + \varepsilon} \;\times\; \frac{b_2 + b_3}{b_1 + b_5 + \varepsilon}
= -\frac{\text{Blue} - \text{Red} + \text{RE}}{{\text{Blue} + \text{Red} + \text{RE} + \varepsilon}} \;\times\; \frac{\text{Green} + \text{Red}}{\text{Blue} + \text{NIR} + \varepsilon}
\end{equation}
The first factor is an ND3 term that is large and positive when the red-edge ($b_4$) is elevated and red ($b_3$) is suppressed. At early establishment, wheat canopy achieves the highest ND3 value (0.757) owing to its high red-edge reflectance ($b_4 = 0.211$) and low red reflectance ($b_3 = 0.035$), while soil scores lowest (0.266) because its high red reflectance ($b_3 = 0.123$) pulls the numerator down. The second factor, SR2, is an asymmetric ratio of visible-band reflectance to the blue-plus-NIR sum; it is smallest when NIR is large. Wheat canopy's high NIR reflectance ($b_5 = 0.665$) yields the smallest 
SR2 value (0.177), whereas soil with low NIR ($b_5 = 0.170$) produces a 
large SR2 (0.903). Because SR2 varies much more across classes (0.177 to 
0.903) than ND3 (0.266 to 0.757), the product is dominated by the SR2 
term, pulling the canopy product to 0.133 — lower than non-wheat 
vegetation (0.252) and soil (0.239).After negation, wheat canopy achieves the largest (least negative) WCI$_1$ value ($-0.133$). Although all classes produce negative index values, canopy values lie closest to zero, so a simple threshold on the index separates canopy from background.

\textbf{Growth Stage~2 index} (Zadoks 59 -- mid season):
\begin{equation}
\label{eq:wci_t2}
\text{WCI}_{2} = -\frac{b_1 + b_2 + b_4}{b_3 + \varepsilon} \;\times\; \frac{b_1 + b_3 + b_4}{b_5 + \varepsilon}
= -\frac{\text{Blue} + \text{Green} + \text{RE}}{\text{Red} + \varepsilon} \;\times\; \frac{\text{Blue} + \text{Red} + \text{RE}}{\text{NIR} + \varepsilon}
\end{equation}
Both factors are SR3 ratios. The first, $(b_1+b_2+b_4)/b_3$, is amplified by a high 
red-edge and low red; the second, $(b_1+b_3+b_4)/b_5$, is suppressed by a large NIR 
denominator. Notably, at mid-season non-wheat vegetation scores slightly higher than 
canopy on the first SR3 term (9.17 vs.\ 8.42) because its red-edge-to-red ratio is 
favorable. The decisive factor is the second SR3 term: at full canopy closure, wheat 
NIR reaches $b_5 = 0.658$, giving a SR3b value of only 0.365, compared to 0.544 for 
non-wheat vegetation ($b_5 = 0.587$) and 2.04 for soil ($b_5 = 0.245$). Although 
non-wheat vegetation has a marginally larger first SR3 factor, canopy's substantially 
lower second factor means the overall product is smallest for canopy. The negation therefore places wheat at the largest WCI$_2$ values (closest to zero), even though all classes produce negative index values. The selection of SR3 over 
normalized differences is consistent with asymmetric ratios providing stronger 
separation when NIR contrast dominates.

\textbf{Growth Stage~3 index} (Zadoks 87 -- late season):
\begin{eqnarray}
\label{eq:wci_t3}
\text{WCI}_{3} &=& \frac{b_1 - b_3}{b_1 + b_3 + \varepsilon} \;\times\; \tanh\!\left(b_3 - \max(b_1,b_2,b_4,b_5)\right) \nonumber \\
&=& \frac{\text{Blue} - \text{Red}}{\text{Blue} + \text{Red} + \varepsilon} \;\times\; \tanh\!\left(\text{Red} - \max(\text{Blue},\text{Green},\text{RE},\text{NIR})\right)
\end{eqnarray}
By Growth Stage~3, the spectral conditions have reversed: wheat canopy has begun to 
senesce, with red reflectance rising to $b_3 = 0.187$ and NIR declining to $b_5 = 
0.455$, while non-wheat vegetation remains actively green with very low red ($b_3 = 
0.054$) and very high NIR ($b_5 = 0.683$). The first factor, $\mathrm{ND}(b_1, b_3)$, 
is strongly negative for canopy ($-0.428$) because red now substantially exceeds blue 
in absolute reflectance ($b_3 = 0.187$ vs.\ $b_1 = 0.075$). For non-wheat vegetation, 
blue and red are nearly equal in magnitude ($b_1 \approx b_3 \approx 0.05$), collapsing 
the ND term to near zero ($-0.038$). The second factor, $\mathrm{HAbs}(b_3)$, is the 
$\tanh$-bounded gap between $b_3$ and the spectral maximum of all other bands; it is 
negative for all classes because $b_3$ is never the spectral peak. For canopy this gap 
is moderate ($b_3 - b_5 = -0.268$, HAbs $= -0.261$), whereas for non-wheat vegetation 
the gap is much larger ($b_3 - b_5 = -0.629$, HAbs $= -0.556$). 
Canopy therefore achieves the largest WCI$_3$ value ($+0.107$), because both factors are moderately negative, producing a positive product. In contrast, the ND term for non-wheat vegetation collapses toward zero, yielding a much smaller product $\approx 0.006$. This formulation also avoids NIR entirely, which is consistent with the spectral reversal described above.

\begin{figure}[htbp]
    \centering
    \includegraphics[width=1\textwidth]{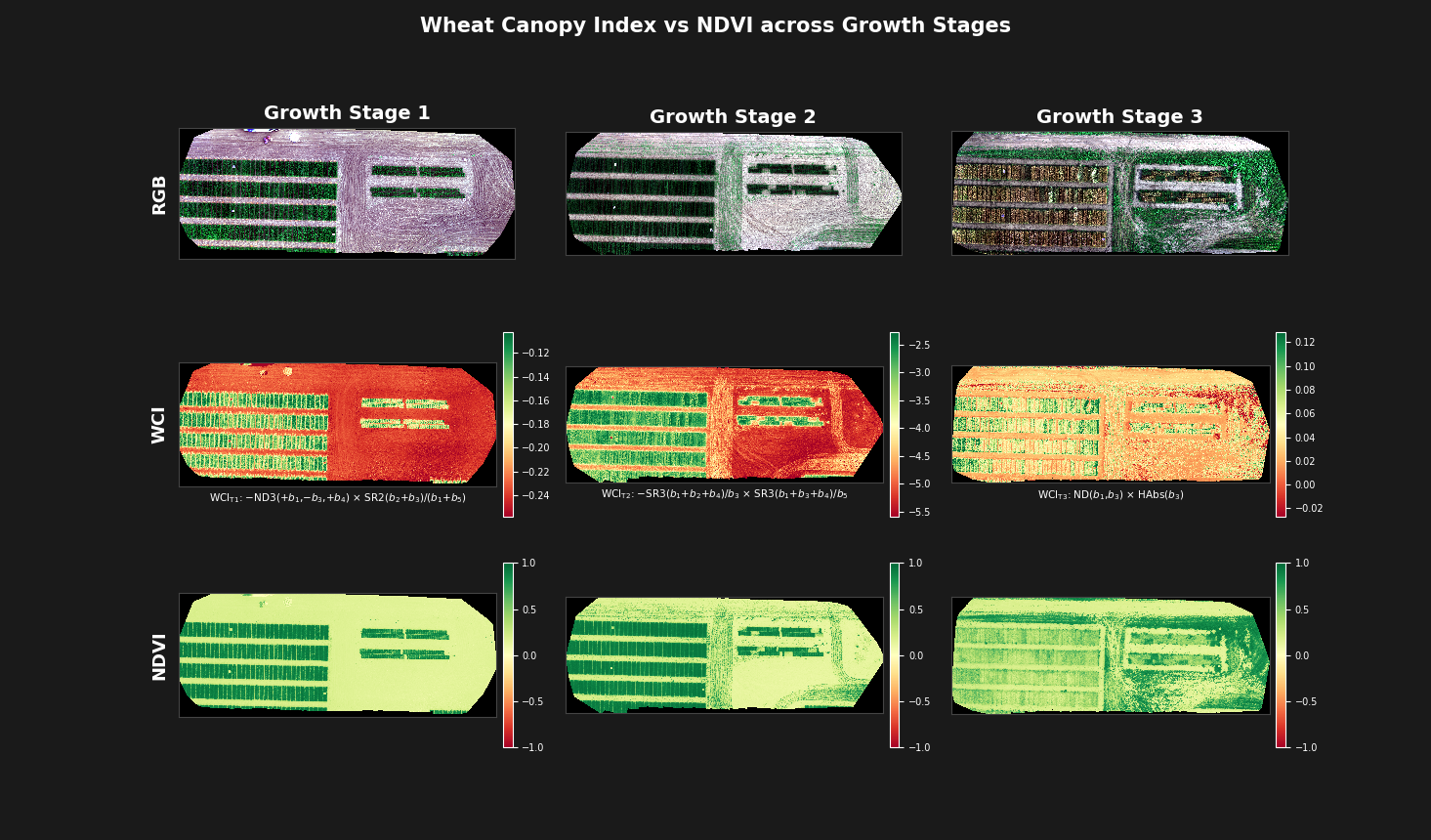}
    \caption{Spatial comparison of RGB, WCI, and NDVI across three growth stages. The WCI produces a clearer canopy--background contrast than NDVI at all stages, with the advantage most pronounced at Growth Stage~3 when conventional spectral contrasts fail.}
    \label{fig:wci_spatial}
\end{figure}

Figure~\ref{fig:wci_spatial} illustrates the spatial behavior of the three WCI formulations across growth stages, compared against NDVI and a true-color RGB composite. At all three stages, the WCI produces a stronger visual contrast between wheat canopy and non-canopy areas than NDVI, with the advantage most pronounced at Growth Stage~3 when conventional spectral contrasts fail.

\subsection{Comparison with Established Vegetation Indices}

Established vegetation indices exhibit two distinct failure modes when applied across 
wheat growth stages: early decline, where indices dependent on NIR--red contrasts 
degrade as canopy closes and soil background diminishes, and late-season loss of discriminatory power, where 
red-edge and NIR indices that perform well at Growth Stage~1 and Growth Stage~2 lose 
discriminative power at Growth Stage~3 as wheat canopy senesces. At late season, NIR reflectance actually reverses order; non-wheat vegetation ($b_5 = 0.683$) exceeds 
senescing canopy ($b_5 = 0.455$), so indices that reward high NIR rank non-wheat 
vegetation above canopy. The WCI avoids both failure modes by selecting a different 
formulation for each growth stage.

We evaluated nine established vegetation indices under the identical spatial block CV 
protocol, using the same linear SVM classifier and $C$ value as each corresponding 
WCI. Table~\ref{tab:uav_comparison} reports mean accuracy and the spatial consistency 
gap (Gap $= |$Block1$\to$2 $-$ Block2$\to$1$|$, in percentage points) for each index 
at each growth stage; a large gap indicates that the index learned a decision threshold 
in one spatial region of the field that fails to transfer to the other.

\begin{table}[htbp]
\centering
\caption{Mean classification accuracy (\%) and spatial consistency gap (Gap $= |$Block1$\to$2 $-$ Block2$\to$1$|$, in percentage points) for the WCI and established vegetation indices under spatial block cross-validation. A large Gap indicates spatial inconsistency: the SVM threshold learned from one block does not transfer to the other.}
\label{tab:uav_comparison}
\begin{tabular}{lcccccc}
\toprule
 & \multicolumn{2}{c}{Stage 1} & \multicolumn{2}{c}{Stage 2} & \multicolumn{2}{c}{Stage 3} \\
\cmidrule(lr){2-3}\cmidrule(lr){4-5}\cmidrule(lr){6-7}
Index & Mean & Gap & Mean & Gap & Mean & Gap \\
\midrule
\textbf{WCI} & \textbf{99.51} & 0.66 & \textbf{97.20} & 2.85 & \textbf{93.54} & 3.28 \\
CIre   & 98.83 & 0.64 & 94.38 & 5.01 & 78.28 & 4.62 \\
NDRE   & 97.88 & 3.24 & 94.17 & 6.12 & 78.01 & 4.09 \\
SR     & 97.19 & 0.34 & 73.12 & 7.44 & 82.58 & 2.96 \\
GNDVI  & 95.59 & 8.13 & 92.45 & 7.94 & 62.98 & 24.71 \\
NDWI\tablefootnote{NDWI and GNDVI reduce to the same Green/NIR ratio under the five-band MicaSense sensor, which lacks a SWIR band. Both are retained in the table for completeness.}   & 95.59 & 8.13 & 92.45 & 7.94 & 62.98 & 24.71 \\
EVI2   & 87.53 & 6.00 & 74.26 & 8.63 & 77.75 & 2.35 \\
SAVI   & 85.87 & 5.17 & 74.46 & 9.01 & 78.57 & 2.92 \\
GRRI   & 77.43 & 19.53 & 55.10 & 5.22 & 83.73 & 4.82 \\
NDVI   & 81.67 & 35.66 & 75.24 & 20.39 & 81.76 & 4.09 \\
\bottomrule
\end{tabular}
\end{table}

The WCI outperformed all established indices at every growth stage by a margin that widens as the season advances: the gap over the second-best index grows from 0.68 percentage points at Growth Stage~1 to 2.82 at Growth Stage~2 and 9.81 at Growth Stage~3. Indices that perform competitively at Growth Stage~1, CIre (98.83\%) and NDRE (97.88\%), lose discriminatory power at late season, degrading to 78.28\% and 78.01\% respectively by Growth Stage~3, while the WCI sustains 93.54\%. Indices such as NDVI and SAVI illustrate the early decline mode, losing substantial accuracy between Growth Stage~1 and Growth Stage~2 as the spectral contrast they rely on weakens with advancing canopy development. SR exhibits the most severe early decline of any index, losing 24.07 percentage points between Growth Stage~1 and Growth Stage~2 (97.19\% to 73.12\%), far exceeding NDVI's 6.43-point drop over the same period, suggesting that pure band ratios are especially sensitive to the changing soil background as canopy closes. GNDVI and NDWI show the most dramatic loss of discriminatory power at late season, falling to 62.98\% at Growth Stage~3 with a 24.71-point spatial gap, failing to generalize across either time or space simultaneously.

\begin{figure}[htbp]
    \centering
    \includegraphics[width=\textwidth]{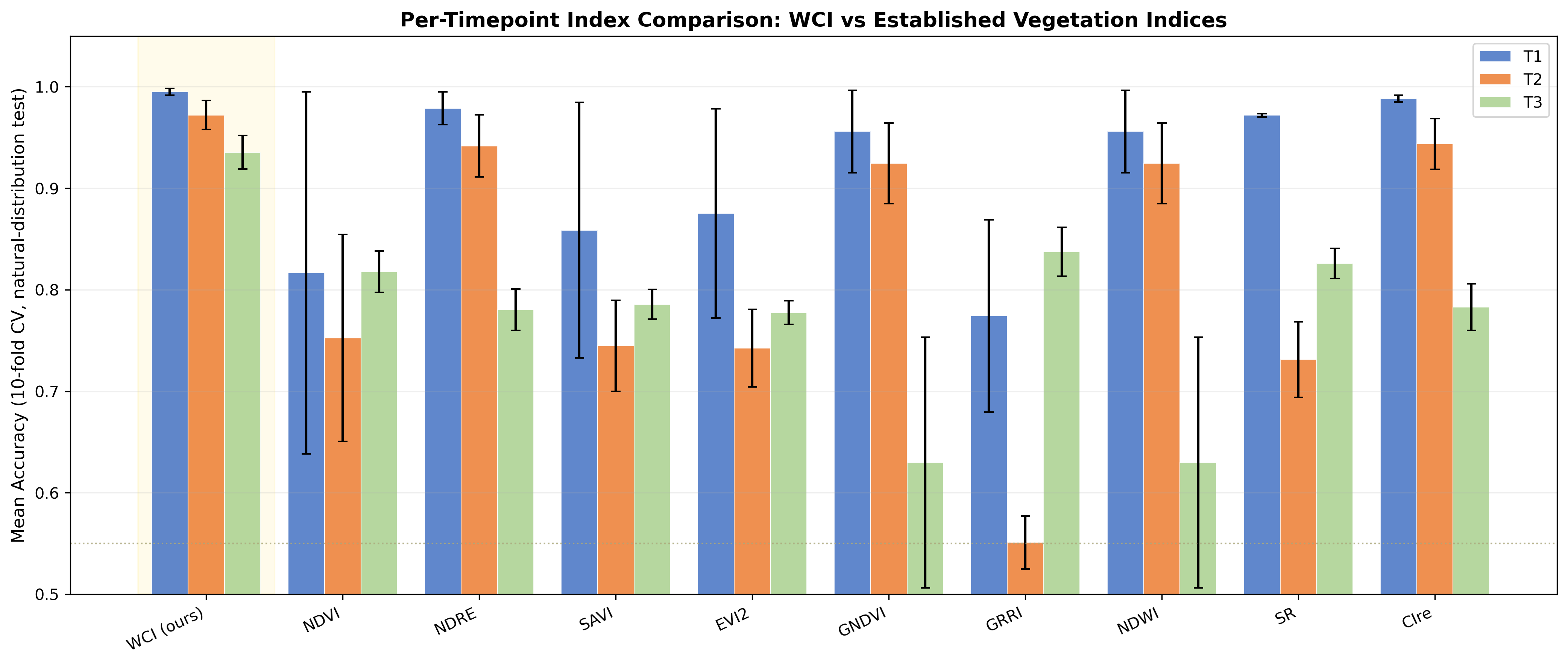}
    \caption{Spatial block CV accuracy for the WCI and nine established vegetation indices across three growth stages. Each bar represents mean accuracy over two spatially separated folds. Two failure modes are visible: early decline (NDVI, SAVI, SR) and loss of discriminatory power at late season (NDRE, CIre, GNDVI).}
    \label{fig:uav_comparison_chart}
\end{figure}

The spatial consistency gap further reveals a pattern invisible in mean accuracy alone. NDVI shows a 35.66-point gap between blocks at Growth Stage~1, indicating that its SVM decision threshold learned from one spatial region of the field does not transfer to the other. The WCI gap at Growth Stage~1 is 0.66 points. Figure~\ref{fig:uav_comparison_chart} summarizes the full cross-stage performance trajectory for all indices.

\section{Discussion}\label{discussion}

The SFP framework occupies a distinct position among automated index discovery methods. Rather than searching an unconstrained space of arbitrary mathematical expressions, SFP restricts the search to ratio-based formulations that preserve the properties that make classical indices reliable across sensors and acquisition conditions—illumination invariance, bounded output range, and normalization by construction. Within this constrained family, SFP performs exhaustive enumeration followed by cross-validated selection to identify the best-performing equation. The result is a reproducible pipeline whose discovered indices inherit the invariance and normalization properties of established indices such as NDVI.

A second distinction concerns the role of the classifier. In many automated discovery pipelines, the index form and the decision boundary are learned jointly. In SFP, the index and the classifier are decoupled: the index is discovered first, and the SVM decision threshold is derived afterwards from the index values. This decoupling means that once discovered, the index requires no model artifacts, standardization statistics, or access to the original training dataset—only the formula itself and a single decision threshold.

The two applications also reveal that spatial generalization is a by-product of the consensus-based selection protocol rather than an explicit design goal. Because an index must emerge independently across folds with different spatial compositions, features whose discriminative signal is location-specific are naturally filtered out—a useful property for operational deployment where calibration data may cover only part of the target region.

One limitation of the current framework is that coefficient optimization was demonstrated only for the Kochia application. The UAV wheat indices were reported without weight tuning, which we treated as a conservative baseline to demonstrate that feature selection alone is sufficient to outperform established indices. Applying coefficient optimization to the WCI formulations is a natural next step that may improve late-season performance at Growth Stage~3, where the margin over the second-best index, while substantial, is narrower than at earlier stages.

\section*{Conclusion}\label{conclusion}
We introduced the Spectral Feature Polynomial (SFP) framework, a general pipeline for 
automatically discovering compact, interpretable spectral indices from multispectral 
imagery. The framework constructs a feature space from thirteen families of spectral 
functions, applies degree-2 polynomial expansion, and uses a combination of ANOVA 
pre-filtering, cross-validated feature selection, and Nelder-Mead coefficient 
optimization to identify a single deployable equation per task. The resulting indices 
run directly on reflectance values and require no standardization statistics at 
deployment time, making them suitable for operational use on any remote sensing 
platform including Google Earth Engine.

We demonstrated the framework on two agricultural applications. For Kochia detection in Sentinel-2 imagery across three growing seasons, the framework converged on the same two-term equation in 44 of 46 independent cross-validation folds, achieving 98.6\% mean accuracy across all 46 folds and outperforming the best established index by more than 4 percentage points under year-held-out evaluation. Coefficient optimization contributed a consistent 1.1--1.5 percentage point improvement across all cross-validation strategies. For wheat canopy classification in UAV multispectral imagery, the framework discovered stage-specific indices that achieved 99.5\%, 97.2\%, and 93.5\% accuracy at Growth Stages 1, 2, and 3, respectively, compared to 78\% or below for the best established index at late season. Critically, the Stage 3 index avoided NIR entirely and exploited blue-red contrast instead, correctly adapting to the spectral reversal that occurs as wheat senesces while non-wheat vegetation remains 
green, a failure mode that caused all nine established indices to lose discriminatory power at late season.

These results suggest that automated spectral index discovery, when grounded in 
interpretable ratio-based formulations and rigorously cross-validated, can produce 
operationally deployable indices that outperform manually designed alternatives while remaining transparent to domain experts. Future work may extend the SFP framework to hyperspectral sensors, regression tasks such as biomass or yield estimation, and multi-temporal index discovery where the optimal formula is allowed to evolve across acquisition dates. The ndindex Python package implementing the SFP framework and all scripts used in this study will be made publicly available upon acceptance.

\section*{Declaration of competing interest}
The authors declare that they have no known competing financial interests or personal relationships that could have appeared to influence the work reported in this paper.

\section*{Acknowledgments}
This work was supported by the Saskatchewan Ministry of Agriculture through the Agriculture Development Fund [grant number 20230164].
\bibliographystyle{plain}
\bibliography{bibfile}
\end{document}